\def\year{2022}\relax
\documentclass[letterpaper]{article} 
\usepackage{aaai22}  
\usepackage{times}  
\usepackage{helvet}  
\usepackage{courier}  
\usepackage[hyphens]{url}  
\usepackage{graphicx} 
\usepackage{natbib}  
\usepackage{caption} 
\DeclareCaptionStyle{ruled}{labelfont=normalfont,labelsep=colon,strut=off} 
\usepackage{algorithm}
\usepackage[noend]{algpseudocode} 

\usepackage[dvipsnames]{xcolor}

\usepackage{newfloat}
\usepackage{listings}
\floatstyle{ruled}
\newfloat{listing}{tb}{lst}{}
\floatname{listing}{Listing}
\usepackage{hhline} 
\usepackage{amsmath}
\usepackage{multirow}
\usepackage{placeins}
\usepackage{amsthm}
\usepackage{amssymb}

\title{Learning Local Heuristics for Search-Based Navigation Planning}
\author{
    Rishi Veerapaneni, Muhammad Suhail Saleem, Maxim Likhachev
}
\affiliations{
    Robotics Institute, Carnegie Mellon University \\
    \{rveerapa, msaleem2, mlikhach\}@andrew.cmu.edu
}

\begin{document}

\maketitle

\begin{abstract}
Graph search planning algorithms for navigation typically rely heavily on heuristics to efficiently plan paths. As a result, while such approaches require no training phase and can directly plan long horizon paths, they often require careful hand designing of informative heuristic functions.
Recent works have started bypassing hand designed heuristics by using machine learning to learn heuristic functions that guide the search algorithm. 
While these methods can learn complex heuristic functions from raw input, they i) require a significant training phase and ii) do not generalize well to new maps and longer horizon paths. 
Our contribution is showing that instead of learning a global heuristic estimate, we can define and learn local heuristics which results in a significantly smaller learning problem and improves generalization. We show that using such local heuristics can reduce node expansions by 2-20x while maintaining bounded suboptimality, are easy to train, and generalize to new maps \& long horizon plans. 
\end{abstract}

\section{Introduction}\label{sec:intro}
Motion planning has many applications like autonomous car navigation, robotic arm manipulation, and multi-agent warehouse autonomy. 
Graph search is one popular class of motion planning methods which relies on typically hand-designed informative heuristics (cost-to-go estimates) for competitive performance \shortcite{urbanChallenge2008,mha2014,improvedMHA,wdg2019}.

The majority of graph search algorithms assume known environmental knowledge (i.e. the graph) to compute valid/invalid nodes and edges for the search algorithm. Given this transition information, heuristic search algorithms can directly work on maps without any computationally expensive pre-training phase. These methods can also solve long horizon tasks out-of-the-box without any algorithmic changes. Additionally heuristic search algorithms have strong theoretical guarantees of completeness and bounded suboptimality given enough computation time. 

Modern machine learning techniques, e.g. reinforcement or imitation learning, on the other hand utilize observational data from the environment to determine paths. These methods bypass needing hand-crafted heuristics by learning complex values and/or policies directly from raw input and perform well on environments similar to those seen during training. 
Recent work have started to bridge the gap of heuristic search and machine learning, usually by learning a neural network which returns a heuristic or priority that is used in a search algorithm (typically weighted A*). These methods show improvements in reducing nodes expanded compared to heuristic search and improvements in increasing success rate compared to pure machine learning (\citeyear{sail2017,deepCubeA2019,learningHeuristicA2020,deepCubeAQ2021,optimalSearchNN}). 
However, most methods require a significant training phase, lack guarantees on completeness or solution quality, and struggle to generalize to long horizon tasks or new maps. 

\begin{figure}
    \centering
    \includegraphics[width=0.45\textwidth]{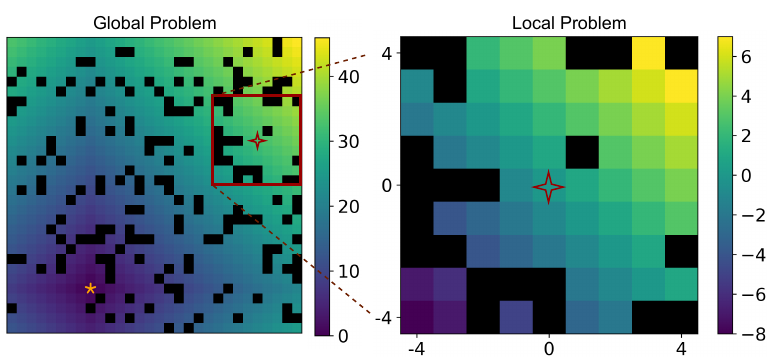}
    \vspace{-2pt}
    \caption{{\footnotesize Left: Estimating a \textit{global} cost-to-go heuristic from the red star state $s$ to the orange goal requires reasoning  over a large region. Right: We define and learn a \textit{local} heuristic centered at $s$ which reasons about vehicle dynamics and obstacles in a local region, that we then combine with the global heuristic. 
    This significantly smaller problem eases the learning progress, enables generalization, and provides significant improvements in some domains.}}
    \label{fig:localglobal}
    \vspace{-8pt}
\end{figure}

\begin{figure}
    \centering
    \includegraphics[width=0.48\textwidth]{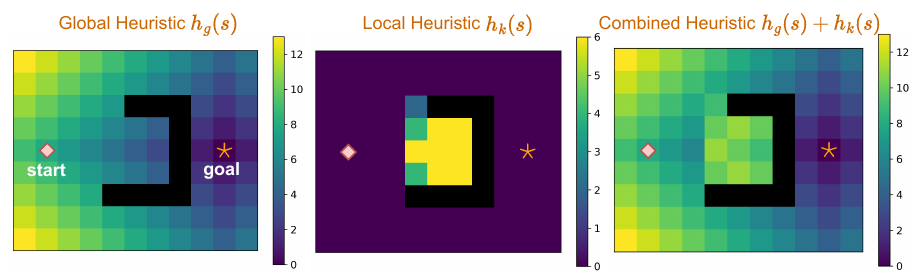}
    \caption{{\footnotesize We demonstrate the effect of computing a local heuristic and summing it to get a combined more informed heuristic. The global heuristic $h_g$ is the Manhattan distance to the goal ignoring obstacles while the local heuristic reasons about obstacles within a window of $K=3$ and inflates heuristic values in the cul-de-sac. Running weighted A* from the start with $h_g+h_k$ now skips the cul-de-sac compared to running with just $h_g$.}
    \label{fig:localglobalexample}}
    \vspace{-10pt}
\end{figure}

Our goal is to combine the best of two worlds of heuristic search's ability to solve long-horizon tasks and machine learning's ability to incorporate environmental data. Concretely, we would like a method that
1. Uses environment data to speed up performance
2. Is easy to train
3. Generalizes to new maps and long horizon plans
4. Maintains solution completeness and suboptimality guarantees.
Our main insight is that instead of learning a global cost-to-go heuristic which becomes exceedingly difficult to train for long horizon plans, we can instead define and learn a \textit{local} heuristic.
Unlike all related literature (to the best of our knowledge) which attempt to directly predict the entire cost-to-goal heuristic, we attempt to only predict the cost to escape a small region centered at the current robot state, enabling easy learning for a neural network (see Figure \ref{fig:localglobal}). 
This local estimate is augmented with the global heuristic to provide a more accurate and informative cost-to-go estimate.
We employ focal search \cite{focalSearch1982}, a variant of A*, to use the informed heuristic in a manner that provides bounded suboptimal guarantees. 
We call this framework Local Heuristic A*, LoHA*, and show how LoHA* can effectively generalize and improve performance.

Succinctly, our main contributions are:
\begin{enumerate}
    \item Defining a \textit{local} heuristic $h_k$ that is \textit{independent} of the full scale planning problem, and using a neural network to estimate its value efficiently.
    \item Combining the local and global heuristic, and using focal search to maintain bounded suboptimality.
    \item Experimentally demonstrating that a learned local 9x9 heuristic can result in 20x node reductions compared to regular weighted A* search on large 1024x1024 maps, and that LoHA* effectively generalizes to new maps. 
\end{enumerate}


\section{Related work}
The majority of prior works incorporating machine learning with search-based planning do so by attempting to directly learn the cost-to-go heuristic to the goal state. \citet{deepCubeA2019,deepCubeAQ2021} learn such functions on a Rubik Cube and other combinatorial tasks (e.g. 24-tile problem, Sokoban) using reinforcement learning.
\citet{learningHeuristicA2020} tests and trains a global heuristic function on \textit{one} map. \citet{optimalSearchNN} impressively, but at the cost of extra machinery, learns an \textit{admissible} CNN heuristic function to find optimal solutions for tile and TopSpin problems. \citet{learningGlobalHF2011} is an early work which uses curriculum learning and a small NN to learn global heuristics on different classical combinatorial problems (e.g. 3x3 Rubik Cube, 24-tile problem). 
For all these methods, it is unclear how their learnt heuristic would work on larger horizon problems or similar but different scenarios outside of their training distribution, e.g. a different goal Rubik Cube state, or similarly-generated but larger maps. Our work aims to speed-up search using machine learning in a way which generalizes to new maps. 

A few other works attempt to speed up search by learning different metrics.
\citet{sail2017} learns a global priority value of features of the search state which determines their expansion policy. 
On the other hand we learn a local heuristic based on local features, which allows our method to be useable across different search instantiations (e.g. different weights).
\citet{learnExpansionDelay2021} learns an expansion delay heuristic that speeds up search but must be retrained for new maps.


We aim to generalize to different maps by limiting the learning problem to a ``local" sub-problem.
Our local sub-problem is loosely related to lookahead in best-first search \cite{lookaheadOptimalBFS} which uses a fixed depth DFS lookahead to update the heuristic in a A* search, but our local heuristic definition is completely different as well as our use of a neural network. Our local definition dramatically eases the learning problem and substantially reduces the required training dataset size, training time, and model size of the neural network while simultaneously enabling it to effectively generalize to different maps.

\section{Method}
Our main motivation is straightforward; simplify the learning problem.
Learning a heuristic which estimates the \textit{global} cost-to-go requires complex reasoning about the entire map. Our main insight is that instead of solving the entire shortest path problem, we define a \textit{local} problem which is significantly easier to solve, and that solving this local problem can result in a large overall reduction of nodes expanded when used in heuristic search.

\subsection{Defining the Local Heuristic}
Heuristic search methods like A* conduct a best first search over states, with their priority $f(s)$ equal to the sum of cost-to-come $g(s)$ and cost-to-go estimate $h_g(s)$. Crucially, $h_g(s)$ (under)estimates the total cost-to-go, i.e. the best cost to reach the goal state $s_g$ from $s$. We call this a \textit{global} heuristic $h_g(s)$, distinct from our local heuristic $h_k(s)$. This means that as the planning problem gets longer/larger, obtaining accurate $h_g(s)$ estimates becomes harder. 
 
We instead propose to learn a local heuristic $h_{k}(s)$ that takes full consideration of the robot dynamics and environmental obstacles in a local region of size $K$ around $s$.
Conceptually, $h_k$ tries to predict the additional cost required to escape the local region. We can then use $h_{gk}(s) = h_g(s) + h_k(s)$ during search (see Figure \ref{fig:localglobalexample}).

Mathematically, given a state $s = (x,y,\Omega)$ with position $x,y$ and other state parameters $\Omega$ (e.g. heading, velocity), we define a local region $LR(s)$ to contain the states within a window of $K$, i.e. $LR(s) = \{s' \mid K \geq |s.x-s'.x|, K \geq |s.y-s'.y|\}$. Let $LRB(s)$ be the border of this region, i.e. $\{s' \mid K = |s.x-s'.x| \vee K = |s.y-s'.y| \}$. 
Conceptually, assuming unit length actions, any path from $s$ to $s_g$ must contain a state in $LRB(s)$, or directly reach the goal in the local region $LR(s)$. If neither are possible from $s$, then $s$ cannot leave $LR(s)$, is in a dead end, and should have an infinite heuristic value. Thus our objective value $h_{gk}(s)$ is 
\begin{equation} \label{equation:localH}
h_{gk}(s) = \min_{s'}
\begin{cases}
    c(s,s')+h_g(s'), & s' \in LRB(s) \\
    c(s,s')+0, & s' = s_g \in LR(s)\\
    \infty, & \text{otherwise}
\end{cases}
\end{equation}

Notice how computing $c(s,s')$, the minimum cost of a path from $s$ to $s'$, requires incorporating the robot's dynamics/kinematic constraints as well as local obstacle/environmental data in $LR(s)$. 
We can compute $h_{gk}(s)$ at a given state $s$ by running A* following Equation \ref{equation:localH}, however this becomes slow as the size of $LR(s)$ increases. We can instead approximate this value by training a neural network (NN). We can input $s$, the environment's data in $LR(s)$, and the heuristic data in $LR(s)$, and predict $h_{gk}(s)$.

A key problem with this approach is that even though our problem is local, our input $s$ and $h_g(s')$ are not scale-invariant. 
For example, if we trained on small maps, but then evaluated on larger maps, our neural network would be unable to generalize to larger encountered $s$ and $h_g$ values.
A key observation is that we can make our inputs invariant to any such changes. The state $s = (x,y,\Omega)$ can become just $\Omega$ as the local region $LR(s)$ is centered at $x,y$. We remove global dependence on $h_g(s')$ for $s' \in LR(s)$ by subtracting $h_g(s)$. Our local invariant heuristic thus becomes

\begin{equation} \label{equation:localInvariant}
\resizebox{1\hsize}{!}{$
h_k(s) = \min_{s'}
\begin{cases}
    (c(s,s')+h_g(s'))-h_g(s), & s' \in LRB(s) \\
    (c(s,s')+0-h_g(s)), & s' = s_g \in LR(s)\\
    \infty, & \text{otherwise}
\end{cases}
$}
\end{equation}

Therefore instead of passing $h_g$ into the NN, we only need the relative information $h_g(s') - h_g(s) \in LR(s)$.


We can generalize this definition for non-unit length actions by predicting the additional cost required to \textit{escape} $LR(s)$. We omit the mathematical definitions for brevity but note that our experiments uses this more general version.




\subsection{Computing Ground Truth $\mathbf{h_k}$}
Equation \ref{equation:localInvariant} defines a multi-goal search problem within $LR(s)$ where we want to minimize $c(s,s') + h_g(s') - h_g(s)$. We directly run an A* search starting at $s$ until either of the first two conditions are met, or until it returns no solution found which results in the third $\infty$ value. In high dimensional state spaces where the number of states within $LR(s)$ is large, it can take prohibitively long for the local search to terminate. We can ease this by conducting a maximum number of expansions and then returning the top $g(s') + h(s')$ in the queue as this is an underestimate of $h_k(s)$.

\subsection{Training Procedure}
\textbf{Neural network inputs:} As described earlier, we want to feed in a locally invariant version of $s$ and $LR(s)$ into the neural network. $LR(s)$ contains both the obstacle and invariant heuristic values of window $K$ centered at $s$. 

\textbf{Collecting data:} We utilize supervised learning to train a model to learn $h_k$.
A naive approach to collect training data is to randomly sample states $s$. However, this may over sample regions in the state space that are not relevant during runtime and hurt performance. We thus collect training data by running weighted A* with ground truth local heuristic and storing the inputs $s, LR(s)$ and corresponding true value $h_k(s)$ of states $s$ we encounter during search.

\textbf{Neural network output:} Local heuristic value $h_k(s)$.

\textbf{Loss function:} One issue we discovered when training our neural network is that regressing directly to $h_k$ causes issue as the mean square error objective prioritizes samples with larger values, reducing the prediction quality for many lower range values. An effective alternate we found was regressing to $\log(h_k+1)$ which is a measure of relative error but has better statistical properties than relative error or other alternatives \citep{relativeErrorLog}. The $+1$ is numerically required as $h_k$ can equal 0. 
Additionally we chose to regress to $h_k = 2K$ for dead-ends where $h_k = \infty$, which we found to be sufficiently large.

\subsection{Using the Local Heuristic in Search}
We use $h_{gk}(s) = h_g(s) + h_k(s)$ as our heuristic.
Conceptually, $h_k$ augments $h_g$ with local dynamics and obstacle information.
If $h_k(s)$ is computed accurately (e.g. by a local search), $h_{gk}(s)$ is guaranteed to be admissible and can be used in A* while guaranteeing optimality. However, if $h_k$ is learnt, it can be arbitrarily suboptimal. We therefore employ focal search, using $h_g$ as a consistent heuristic in OPEN and $h_{gk}(s)$ as an inadmissible heuristic in FOCAL, guaranteeing that our solution is bounded suboptimal. We call this framework of learning a local heuristic, combining it with the global heuristic, and using it in focal search, Local Heuristic A*, or LoHA* for short.


\section{Local Heuristic Experiments}
We experiment using custom random obstacle maps and 6 city maps from \cite{sturtevant2012benchmarks}, minimizing travel time between start-goal pairs.
We simulate a non-holonomic car with state $(x,y,\theta,v)$ with positions $x,y$ discretized by 0.5, heading $\theta$ discretized by 30 degrees, and velocity $v \in \{-1,0,1,2,3\}$.
The car follows Ackermann dynamic constraints and every state has unit-cost actions of $\Delta v \in \{-1, 0, 1\}$ and steering angle $\in \{-60, -30, 0, 30, 60\}$. Since the max velocity is 3, our $h_g$ heuristic is $L_2(s, s_{goal})/3$. 
Our objective with this set-up is to show how a local heuristic can help in complex state and action spaces as opposed to many existing works combining search and machine learning on 4/8-connected grids. 
We report results for a small local heuristic size of $K=4$. Experiments were run on an Ubuntu 20.04 machine with 32-GB Ram and a 11th Gen Intel Core i7-11800H@2.30GHzx16. 

\subsection{Training}
\subsubsection{Local state input: } 
We input $LR(s)$ as a 2 channel $2K+1$ by $2K+1$ image centered at $(\text{floor}(x),\text{floor}(y))$. The first channel is the binary obstacle map, the second the local invariant heuristic $h_g(s')-h_g(s)$. We additionally input the local invariant state containing $(x-\text{floor}(x),y-\text{floor}(y),\theta,v)$.
\subsubsection{Training data:} We run weighted A* with the local heuristic on random start-goal locations on a set of training maps, and collect data on states we have seen. We use a local heuristic expansion limit of 100 to enable faster data collection. Overall the procedure is fast; with unoptimized C++ code we collect on the order of 5000 examples a second. We train on 200,000 states (which can be collected in minutes). We highlight that this contrasts learning a global heuristic where data collection takes longer as each training example requires solving the entire planning problem. 
\subsubsection{NN architecture:} We apply a convolutional layer to $LR(s)$, flatten out latent vector, append our local invariant state $s$, and apply two intermediate size 100 MLP layers.
\subsubsection{Training time:} We train on 200,000 examples for 100 epochs with a batchsize of 32 on CPU, which takes roughly 20-30 minutes. We did not optimize training speed but again we iterate our local problem enables a smaller model and correspondingly smaller compute requirements (i.e. using a CPU and not a GPU, training in minutes and not hours). After training, our squared relative loss saturates around 0.03, corresponding to about 18\% absolute relative error.

\begin{table}[t]
\centering
\resizebox{0.48\textwidth}{!}{
\begin{tabular}{|c|c|c|c|c|c|c|}
\hline
\multirow{2}{*}{Map Type} & \multirow{2}{*}{Split} & \multirow{2}{*}{Method} & \multicolumn{4}{c|}{Reduction in nodes expanded} \\ \cline{4-7}
 &  &  & $w$2 & $w$8 & $w$32 & $w$128 \\ \hhline{|=|=|=|=|=|=|=|}
\multirow{4}{*}{random20} & \multirow{2}{*}{Train} & A* w/TL & 6.76 & 10.88 & 12.78 & 14.7 \\
 &  & LoHA* & 3.53 & 7.92 & 10.33 & 11.6 \\ \cline{2-7}
 & \multirow{2}{*}{Test} & A* w/TL & 6.6 & 10.42 & 14.45 & 15.75 \\
 &  & LoHA* & 3.57 & 6.94 & 10.46 & 12.67 \\ \hline
\multirow{4}{*}{random30} & \multirow{2}{*}{Train} & A* w/TL  & 12.21 & 26.3 & 40.38 & 44.02 \\
 &  & LoHA* & 2.16 & 12.07 & 18.08 & 20.51 \\ \cline{2-7}
 & \multirow{2}{*}{Test} & A* w/TL & 10.36 & 28.58 & 43.57 & 44.3 \\
 &  & LoHA* & 1.68 & 7.71 & 13.59 & 16.55 \\ \hline
\multirow{4}{*}{Denver\_256} & \multirow{2}{*}{Train} & A* w/TL & 2.43 & 6.45 & 5.92 & 7.13 \\
 &  & LoHA* & 1.22 & 5.15 & 3.98 & 6.37 \\ \cline{2-7}
 & \multirow{2}{*}{Test} & A* w/TL & 4.54 & 16.37 & 30.73 & 29.21 \\
 &  & LoHA* & 1.43 & 8.43 & 28.16 & 30.73 \\ \hline
\end{tabular}}
\caption{{\footnotesize
    \textbf{LoHA* Results |} 
We report the median multiplicative reduction in nodes compared to weighted A*. We see that LoHA* is able to get larger reductions as the weight $w$ increases, and that we are able to effectively able to generalize to different maps. 
}}
\vspace{-8pt}
\label{tab:results}
\end{table}

\subsection{Results}
Table \ref{tab:results} reports the median speed-up across several weighted runs of using A* with $h_{gk}$ using the ground Truth Local heuristic (A* w/TL) and LoHA* using a neural network approximation on both the training and testing maps.
The ``randomN" maps are 1024x1024 maps with N\% randomly generated obstacles, split into 7 training and 3 testing maps. The Denver maps are 256x256 split into 2 training maps and 1 testing map. 
Overall, each training/testing set has about 40/20 individual start-goal pairs correspondingly, with 3 seeds run per configuration.
We report the median reduction in nodes expanded compared to the corresponding weighted A* baseline, e.g. a value of 6.76 means the method expands 6.76 times less nodes than weighted A*.

The ``A* w/TL" results reveal the usefulness of the local heuristic in reducing the total number of nodes expanded, ranging from 2-40x depending on the map and heuristic weight $w$. We see that $h_{gk}$ is more effective when $w$ is larger; this occurs as node expansions for larger $w$ are more likely to occur in local optimas while $h_{gk}$ penalizes these regions more. Additionally, our ability to run A* w/TL informs us of the estimated upper-bound that LoHA* can obtain, and determine regimes where LoHA* would not be effective. 
This capability is useful for practitioners as they can easily determine beforehand if LoHA* will be useful for their domains.

\begin{figure}[t]
    \centering
    \includegraphics[width=0.40\textwidth]{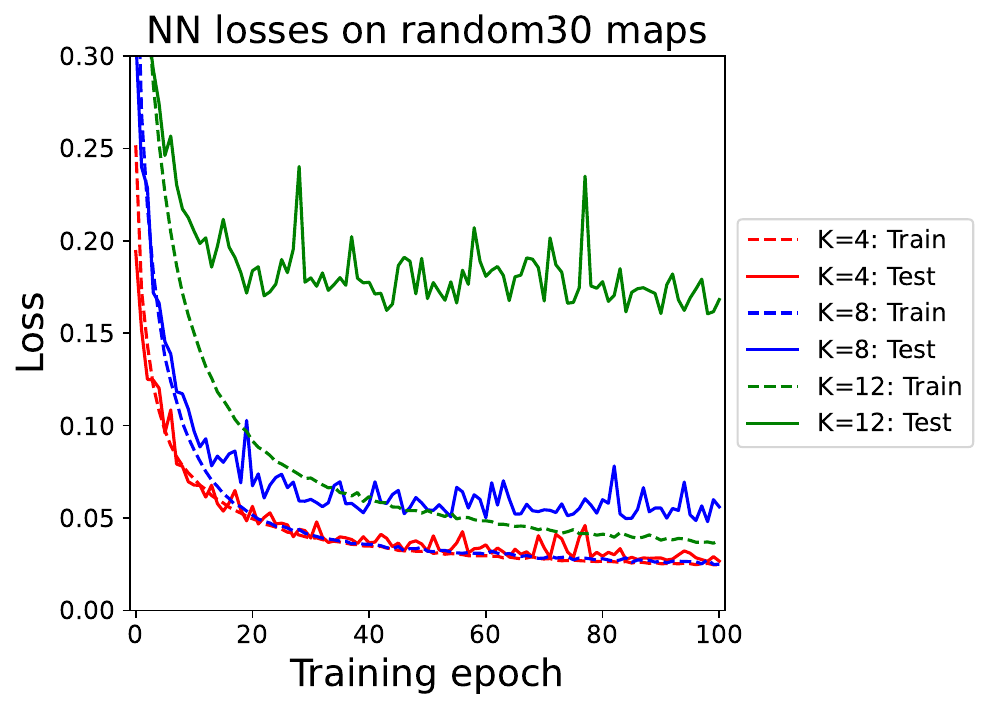}
    \vspace{-8pt}
    \caption{{\footnotesize The y-axis is the log relative loss objective; a loss of 0.2 roughly translates to $\geq$ 50\% absolute relative error, 0.1 to $\geq$ 35\%.  As $K$ increases, the neural network struggles to generalize to the test maps.
    This supports our motivation that learning a local heuristic eases the learning problem and improves generalization.
    }}
    \label{fig:nnloss}
    \vspace{-12pt}
\end{figure}

LoHA* is able to roughly match the order of magnitude of performance of the true local heuristic. We note that some degradation in performance is expected as LoHA*'s neural network is a noisy approximation of the true local heuristic, but see that the noisy approximation is still effective in reducing node expansions. Importantly, LoHA* is able to effectively generalize to the test maps not seen in during training. Figure \ref{fig:nnloss} shows how increasing $K$ makes it harder for the neural network to generalize to testing maps, justifying our motivation for using a local and not global heuristic to enable generalization.

One key limitation with LoHA* is that although it can significantly reduce node expansions, its overall runtime is longer than baseline A*. 
This occurs as running the neural network in the search is slow; LoHA* expands roughly 4,500 nodes a second (with neural network inference time dominating) while A* with $h_g$ expands roughly 140,000 nodes a second. 
We imagine LoHA* will provide runtime benefits in scenarios where node expansions are more expensive, or by utilizing batch expansions in focal search or GPU optimization \cite{kfocal,optimalSearchNN,nonBlockingBatch}. This is independent of our core contribution and is left for the future.

\section{Future Work and Conclusion}
Our key assumption is that we could define a local region around the physical region of the state $s$ of the agent, which works in navigation. Expanding this for other domains, e.g. manipulation, would be interesting future work where defining $LR(s)$ could be non-trivial.
As mentioned in the previous section, future work could also address the runtime issues of using a neural network in a heuristic search loop.

We present a framework for extracting, learning, and using local heuristics in heuristic search in navigation planning. 
Using the local heuristic in a focal A* search results in a significant reduction in nodes expanded compared to regular A*, while maintaining bounded suboptimality gaurantees.
We show that learning a local heuristic enables significantly easier data collection, learning, and generalization while decreasing expansions by 2-20x.
\textbf{Acknowledgements} This material is partially supported by the National Science Foundation Graduate Research Fellowship under Grant No. DGE1745016 and DGE2140739. 

\clearpage
\bibliography{ref} 

\end{document}